\documentclass[default]{sn-jnl}

\jyear{2021}%

\usepackage{amssymb}
\usepackage{pifont}
\usepackage{graphicx}               
\usepackage{amsmath}
\usepackage[dvipsnames]{xcolor}
\usepackage{makecell}
\usepackage{algorithm}
\usepackage{algpseudocode}
\newcommand{\cmark}{\color{ForestGreen} \ding{51}}
\newcommand{\xmark}{\color{red}\ding{55}}

\begin{document}

\title[Data augmentation through multivariate scenario forecasting]{Data augmentation through multivariate scenario forecasting in Data Centers using Generative Adversarial Networks}

\author*[1]{\fnm{Jaime} \sur{Pérez}}\email{jperezs@comillas.edu}

\author[2,3]{\fnm{Patricia} \sur{Arroba}}\email{p.arroba@upm.es}
\equalcont{These authors contributed equally to this work.}

\author[2,3]{\fnm{José M.} \sur{Moya}}\email{jm.moya@upm.es}
\equalcont{These authors contributed equally to this work.}

\affil[1]{IIT - Instituto de Investigación Tecnológica, Universidad Pontificia Comillas, 28015 Madrid, Spain}

\affil[2]{LSI - Integrated System Lab., ETSIT, Universidad Politécnica de Madrid, 28040 Madrid, Spain}

\affil[3]{CCS - Center for Computational Simulation, Campus de Montegancedo UPM, 28660 Madrid, Spain}

\abstract{
The Cloud paradigm is at a critical point in which the existing energy-efficiency techniques are reaching a plateau, while the computing resources demand at Data Center facilities continues to increase exponentially. The main challenge in achieving a global energy efficiency strategy based on Artificial Intelligence is that we need massive amounts of data to feed the algorithms. This paper proposes a time-series data augmentation methodology based on synthetic scenario forecasting within the Data Center. For this purpose, we will implement a powerful generative algorithm: Generative Adversarial Networks (GANs). Specifically, our work combines the disciplines of GAN-based data augmentation and scenario forecasting, filling the gap in the generation of synthetic data in DCs. Furthermore, we propose a methodology to increase the variability and heterogeneity of the generated data by introducing on-demand anomalies without additional effort or expert knowledge. We also suggest the use of Kullback-Leibler Divergence and Mean Squared Error as new metrics in the validation of synthetic time series generation, as they provide a better overall comparison of multivariate data distributions. We validate our approach using real data collected in an operating Data Center, successfully generating synthetic data helpful for prediction and optimization models. Our research will help optimize the energy consumed in Data Centers, although the proposed methodology can be employed in any similar time-series-like problem.
}


\maketitle

\section{Introduction}
\label{sec:intro}

Our society is immersed in an unstoppable digitalization and interconnection process. By 2023, 66\% of the world's population (5.3 billion people) will have access to the Internet, up from 51\% in 2018 (3.9 billion people). Besides, the number of connected devices will increase to more than three times the world's population by 2023 \cite{noauthor_cisco_2020}. This digitization trend brings significant benefits but, at the same time, enormous challenges. Nearly 80\% of global Internet traffic is generated by video streaming, online gaming, and social media applications \cite{noauthor_global_2019}, which base their operations on the Cloud Computing paradigm. This context, together with the demand for novel applications on the Internet of Things (IoT), Artificial Intelligence (AI), and Smart Cities (e.g., eHealth, autonomous vehicles), will lead to an unprecedented amount of data traffic and requests for computing resources in the Cloud. The ever-increasing demands placed on Data Center (DC) facilities will make their energy inefficiencies even more evident and worrisome. It is estimated that cooling energy alone represents, on average, 40\% of the total energy consumed in DCs \cite{team_trends_2017}.

DCs have become the backbone of both the Internet and the telecommunications industry. Between 2010 and 2018, DCs compute instances increased by 550\% and IP traffic by 1,000\%. This increase resulted in global energy use of about 205 TWh in 2018, representing around 1\% of global electricity consumption \cite{masanet_recalibrating_2020}. Most research on the subject agrees that the energy demand of DCs will continue to grow, although the magnitude of this increase is a controversial topic \cite{masanet_recalibrating_2020} \cite{BELKHIR2018448} \cite{jones_how_2018}. This controversy is due to the remarkable improvements in DCs' and servers' energy efficiency, which have led to a 24\% decrease in the overall energy intensity per compute instance, thus leading to exaggerated energy consumption forecasts in many studies \cite{noauthor_intergenerational_2018}. However, we are reaching the limits of current optimization methods. The next doubling of global DC compute instances may occur within the next 3 or 4 years \cite{noauthor_cisco_2018}, and we will require assiduous efforts to handle the likely abrupt growth in energy demand once the existing efficiency resource is fully exploited.

In recent years, the discipline of data-driven optimization has undergone a revolution thanks to Machine Learning (ML) and Deep Learning (DL) techniques. These AI-based algorithms have enabled the implementation of proactive policies that react to real-time changes, giving a considerable competitive advantage to the companies that implement them. Besides, its development is cheaper and less time-consuming than other more classical techniques. On the other hand, the alternative approach to data-driven optimization is explicit physical models. The main limitations of using this more classical approach are: (i) they are more expensive and time-consuming; (ii) cannot quickly adapt to internal or external changes (e.g., weather, physical layout, data anomalies); (iii) when the system under analysis is highly complex, minor errors in the models may lead to considerable errors in the predictions; (iv) require individual and specialized models, making the generalization of solutions impractical. Consequently, more proactive, automatic, and cost-effective alternatives have been sought using data-driven optimizations.

The modern DC is a very complex system, incredibly challenging to optimize due to the sheer number of configurations and nonlinear interdependencies. These facilities involve many mechanical, electrical, and control systems communicating dynamically. The global improvements in Cloud energy efficiency are led by hyperscale DCs, from companies like \textit{Google}, \textit{Amazon}, and \textit{Facebook}. These giant corporations have access to AI-based optimization technologies beyond the reach of more traditional DCs since vast amounts of data and computing power are required. Data-driven science and engineering optimization techniques have enabled hyperscale DCs to work with Power Usage Effectiveness (PUE) of 1.1 or lower (very close to the ideal 1), whereas the global average PUE is 1.59 \cite{noauthor_annual_2020_uptime}. \textit{DeepMind} reported that the use of AI-driven solutions in \textit{Google}'s DCs had yielded cooling energy savings of 40\%, which equates to a 15\% reduction in overall PUE \cite{evans_deepmind_nodate}. They achieved the lowest PUE ever seen in a \textit{Google} facility. These results would be a vast improvement in any large scale energy-consuming environment but, considering how sophisticated \textit{Google}'s DCs are already, it is a phenomenal step forward. 

The main challenge to globally applying this AI-based approach is that we need massive amounts of data to feed the algorithms, especially those based on the DL paradigm. Gathering sufficient data is expensive, time-consuming, and resource-intensive. Furthermore, we can endanger the electronic equipment's integrity by including high variability in the data (i.e., anomalies), necessary for a good model generalization. Companies with access to large amounts of DC data decide not to share them because it can pose a significant security risk. In practice, the benefits of AI-driven optimizations are restricted to those corporations who own the data. Therefore, a fundamental question arises: What can we do if we do not have enough data?

To solve this challenge, AI can help us again through synthetic data generation. Interest in synthetic data has been increasing over the last few years. Companies like \textit{NVIDIA} \cite{lebaredian_synthetic_2019}, \textit{IBM} \cite{duemig_accelerating_2017}, \textit{Google} \cite{kohlberger_generating_2020}, and agencies such as the \textit{US Census Bureau} \cite{noauthor_modernization_2020} have adopted different synthetic data methodologies for improved model building, application development, and data dissemination. This technique provides realistic data efficiently and at scale when access to real data is too costly, dangerous, or unethical. The generated data can improve data-driven models and statistical analysis. Furthermore, it is applied as a data anonymization method, preventing sensitive information leakage and promoting data-sharing between companies and institutions. At a conceptual level, synthetic data are generated from real samples and have similar statistical properties. Real data are complicated and messy, and data synthesis needs to work within that context. The degree to which a synthetic dataset is an accurate approximation for the real one is a standard measure of realism. 

It is well known that for supervised learning-based techniques, training on augmented datasets can increase the performance of the models \cite{wang2017effectiveness} and that too small a training dataset can lead to overfitting. While many dataset augmentation techniques exist for image data \cite{shorten_survey_2019}, these methods do not generalize well to time series. Traditional methods for time-series data augmentation either arbitrarily cut the time series, tending to remove temporal correlation in the data, or do not generalize well in general. For these reasons, we consider appropriate the use of Deep Learning in the generation of synthetic data, and specifically Generative Adversarial Networks (GAN) \cite{goodfellow_generative_2014}, which have proven in recent years to be one of the most successful algorithms for generating high-quality synthetic data (particularly images and videos). In addition, the use of GANs allows us to handle large amounts of multivariate data with complex nonlinear relationships and different natures (e.g., categorical). Furthermore, Z. Li et al. \cite{ZhengTSAGAN2021} empirically demonstrate the use of GANs to generate synthetic time series data increase the performance between 8.3\% to 12.5\%, on 85 datasets of the UCR 2015 archive.

This paper will apply a Conditional WGAN-GP architecture \cite{NIPS2017_892c3b1c} to generate heterogeneous multivariate computing scenarios, thus addressing the proposed problem of synthetic data generation in DCs and some limitations identified in the state of the art. Our proposal also includes a method to generate anomalous situations on demand, increasing the variability of the data so that the DC's operational management can be trained to handle these sporadic situations. Specifically, the main contributions of our work are the following:

\begin{itemize}
    \item A data augmentation methodology for DCs that brings together the disciplines of GAN-based data augmentation and scenario forecasting. This proposal fills the gap in the generation of synthetic data in DCs, enabling the increase in the volume of open and diverse data in this field.
    \item The generation of on-demand anomalous scenarios, which increases the data heterogeneity without additional effort and without compromising the electronic equipment's integrity.
    \item The suggestion to use the Kullback-Leibler Divergence and Mean Squared Error as metrics in the results, providing a more global comparison between the multidimensional distributions of real and synthetic data.
\end{itemize}

The remainder of the paper is organized as follows: Section \ref{sec:literature} gives a comprehensive literature review of the classical and the most innovative methods of time-series data augmentation. Section \ref{sec:method} describes the state-of-the-art improvements implemented in this research to stabilize the GAN training phase and the metrics used to evaluate the quality of the results. Section \ref{sec:case-study} presents the use case through which we validate our proposal. Section \ref{sec:experiments} provides an analysis of the conducted experiments and the results obtained. Finally, in Section \ref{sec:conclusion} presents the main conclusions drawn from this research.

\section{Literature Review}
\label{sec:literature}
There are severe limitations in obtaining massive and highly representative amounts of time-series data. The primary constraint is indeed the time required for their collection. If we wish to gather data from a process during one year, we must gather them for that entire year. This limitation arises because the fundamental characteristic of time-series data is that they exhibit a time dependency in their values. To address this restriction, generating synthetic time-series data has been an open challenge in industry and academia. If successfully achieved, it would dramatically boost the optimization of the most relevant real-world processes.

Nevertheless, generating synthetic time-series data that preserve time dependency is far more challenging than synthesizing other data types such as images or tabular data. Tabular data assumes that a single row stores all the information from a particular event. On the other hand, sequential data has time-sensitive information spread across many rows and columns. The length of these sequences is often variable, and many historical events potentially condition each data point across multiple variables. Therefore, minor errors can propagate throughout the sequence, introducing considerable deviations.

Statistical and ML methods for time-series data generation have been extensively studied. Unfortunately, many of these efforts have resulted in low quality or limited flexibility of the generated data. The proposed models were designed for each specific problem in many cases, thus requiring specialized domain knowledge. We will now examine some fundamental approaches on which the literature on time-series generation is based.

\subsection{Classical Approaches}
The most commonly used traditional methods of time-series data augmentation include Gaussian noise addition, rotation, scaling, warping, and permutation \cite{Terry_Data_2017}. B. K. Iwana et al. \cite{iwana2020empirical} recently conducted an empirical survey comparing the results of classical data augmentation methods for time series classification with neural networks. K. Bandara et al. \cite{BandaraImproving2021} empirically proved that specific classical statistical techniques for time-series data augmentation (e.g., moving block bootstrapping, dynamic time warping barycentric averaging) improve the performance of global forecasting models based on recurrent neural networks. These classical techniques are helpful in many problems and can be used in conjunction with other more complex methods, although the available data severely restrict the variability of the extractable synthetic data.

\subsection{Simulation-based Models}
This field consists of building a model that mimics a real system's behavior based on explicitly defined physical laws and heuristic expressions \cite{yu2019simulation} \cite{fernandez2018score}. If the simulator approaches real-world systems, they provide high-fidelity data. However, in practice, tuning the model parameters is a challenging task, even with data-driven methods. Moreover, executing the simulations is computationally expensive, and it requires redesigning and readjusting the simulator for each DC.

\subsection{Discriminative Machine Learning (ML) based Models}
These are general parametric models, in which the parameters are adjusted in a training phase, generally through gradient descent algorithms, using actual data examples. This approach is usually applied for time-series analysis, classification, and forecasting. However, we can use the obtained predictions as synthetic data to train other ML and DL models.

\subsubsection{Autoregressive (AR) Models}
AR models are linear recurrent stochastic processes, where each point in the time series depends linearly on the previous $n$ steps and a stochastic term. AR generalizations such as Autoregressive Integrated Moving Average (ARIMA) are powerful and extensively used approaches in time-series forecasting \cite{siami2018comparison} \cite{conejo_day-ahead_2005}. However, AR models have a fidelity problem, producing simplistic models that cannot capture nonlinear temporal relationships, and their predictions are highly dependent on the available data.

\subsubsection{Markov Models}
Markov Models are a general approach for modeling dynamic pseudo-stochastic categorical systems. They assume the Markov property, meaning that future states depend only on the current state rather than past events. These models are widely adopted in predictive modeling and probabilistic forecasting fields. Variants such as Hidden Markov Models have been applied in time series distributions modeling \cite{zucchini_hidden_2017}. Like AR models, the major limitation of these methods is that they cannot correctly capture nonlinear temporal relationships, and their predictions are highly dependent on the available data.

\subsubsection{Bayesian Models}
Bayesian statistics are widely used in time series modeling \cite{pole2018applied} \cite{west_bayesian_2020} and are ideal for noting an event and predicting that any one of several known possible causes was the contributing factor. For example, it could represent the probabilistic relationships between diseases and symptoms. Dynamic Bayesian Networks are a particular class of Bayesian networks that can perform sequential data modeling. Other thriving models include the Bayesian Structural Time Series (BSTS) and Bayesian Model Averaging. The main limitation of these approaches involves the manual definition of prior probabilities and cannot correctly capture complex long-term relationships. Besides, their predictions are highly dependent on the available data. 

\subsubsection{Recurrent Neural Networks (RNNs)}
RNNs are a type of Artificial Neural Networks (ANNs) in which the connections between neurons incorporate feedback loops and internal states (memory), thus exhibiting a dynamic temporal behavior. The best-known example are the Long Short-Term Memory networks (LSTMs). In recent years, they have been one of the pillars of progress in Artificial Intelligence and have been used with great success in Natural Language Processing, Computer Vision, finance, and prediction of all kinds of complex multivariate time series \cite{perez2019predictive} \cite{yin_deep_2021} \cite{perez2018thermal}, including multirate time series \cite{Huang2021_fault}. However, due to limitations in training methods and the amount of data available for many applications, RNNs have shown substantial limitations. Moreover, like all other methods based on discriminative ML, their predictions highly depend on the available data.

\subsection{Generative Adversarial Networks}
\label{subsec:gan_soa}
In recent years, methods based on Generative Adversarial Networks (GAN) have emerged as a popular technique for augmenting datasets, with outstanding results in images and videos. Unlike more classical statistical methods, neural networks can handle multivariate data with nonlinear relationships from different natures (e.g., categorical data, text, etc.). Due to their nature as generative algorithms, GANs may understand better the complex nature of time-series data. Models from the discriminative ML paradigm attempt to learn the conditional probability distribution of the labels $Y$ given the observations $x$, symbolically expressed as $P(Y\mid X=x)$. On the other hand, generative models such as GANs directly estimate the conditional probability distribution of the data $X$ given the observations $x$, symbolically expressed as $P(X=x)$. 

GANs have been proposed in some research for time-series data augmentation. The ground-breaking publication was presented by C. Esteban et al. \cite{esteban_real-valued_2017}. The authors produce realistic-looking heterogeneous medical time-series data using a Recurrent Conditional GAN architecture (RCGAN). Related solutions to generate synthetic data have been successfully applied in the energy sector \cite{lan_demand_2018} \cite{Fekri_Generating_2020} \cite{zhang_generative_2018}, sensory data \cite{alzantot_sensegen_2017}, and health applications \cite{alharbi_synthetic_2020} \cite{norgaard_synthetic_2018} \cite{harada_biosignal_2019}. There are also remarkable works in the literature that address the issue of noisy time series or missing data \cite{ZhengTSAGAN2021} \cite{ramponi2018t}. The state-of-the-art algorithm so far in the augmentation of synthetic time-series data was \textit{TimeGAN} \cite{yoon_time-series_2019}. Recent valuable research has addressed some limitations of the \textit{TimeGAN} approach, such as that of H. Ni et al. \cite{ni2020conditional}, where they propose a new metric to simplify GAN training. Another crucial example is Z. Lin et al. \cite{lin_using_2020} (\textit{DoppelGANger}), where authors explore augmenting time-series datasets related to communication network systems with outstanding results.

The main limitation of these proposals is that their objective is solely to obtain similarity in their predictions, i.e., to generate synthetic data almost indistinguishable from real data. We believe they could benefit significantly by taking advantage of the flexibility offered by generative algorithms such as GANs. Therefore, our work focuses on merging the data augmentation approach with scenario forecasting to increase the exploitation of available real data to extract synthetic data.

Regarding scenario forecasting methods, some publications address it through classical statistical methods \cite{wang_scenario_2017}, although these approaches require extensive knowledge of the specific problem and have severe limitations in the quality and flexibility of the obtained results. The emergence of GANs has provided a renewed impulse to this field. In 2018, Y. Chen et al. \cite{chen_unsupervised_2018} proposed a novel data-driven scenario forecasting approach of energy systems (e.g., wind, solar, load). This model-free approach produces a set of realistic scenarios based on historical data observations. To the best of our knowledge, this approach has only been used in the energy sector \cite{chen_unsupervised_2018} \cite{jiang_scenario_2018} \cite{zhang_typical_2020}. The main difference between these researches and ours is that they aim to generate realistic scenarios but do not intend to use them as synthetic data. In contrast, our work benefits from the variability offered by GANs to generate large amounts of synthetic data. Furthermore, all these proposals operate only with univariate data, and just one proposal \cite{zhang_typical_2020} handles data of different nature (e.g., categorical data).

\begin{table}
\centering
\tiny
\caption{State of the art comparison in synthetic time-series data augmentation and scenario generation using GANs}
\label{tab:soa-gan}
\begin{tabular}{ccccccc}
\toprule
\multirow{2}{*}{Research Features}                                        & \textit{TimeGAN}                       & \textit{DoppelGANger}           & \multicolumn{3}{c}{Scenario Generation}                                             & \multirow{2}{*}{\textit{Ours}}  \\
                                                                          & \cite{yoon_time-series_2019} & \cite{lin_using_2020} & \cite{chen_unsupervised_2018} & \cite{jiang_scenario_2018} & \cite{zhang_typical_2020} &                                 \\ 
\midrule
Data Augmentation                                                         &  \cmark  &  \cmark  &  \xmark  &  \xmark  &  \xmark  &  \cmark  \\
Scenario Generation                                                       &  \xmark  &  \xmark  &  \cmark  &  \cmark  &  \cmark  &  \cmark  \\
\begin{tabular}[c]{@{}c@{}}Generation from any\\time instant\end{tabular} &  \xmark  &  \xmark  &  \cmark  &  \xmark  &  \xmark  &  \cmark  \\
Multivariate Generation                                                   &  \cmark  &  \cmark  &  \xmark  &  \xmark  &  \xmark  &  \cmark  \\
Categorical Variables                                                     &  \xmark  &  \cmark  &  \xmark  &  \xmark  &  \cmark  &  \cmark  \\
On-demand anomalies                                                       &  \xmark  &  \xmark  &  \xmark  &  \xmark  &  \xmark  &  \cmark  \\ 
\bottomrule                            
\end{tabular} 
\end{table}

Our work combines GAN-based data augmentation and scenario generation approaches to generate a large amount of synthetic data, increasing the exploitation of available real data. GAN-based data augmentation approaches aim only to obtain synthetic data similar to real data and do not use the power of control in the generation that GANs offer. In contrast, our research uses the control flexibility offered by GANs to generate numerous realistic scenarios to increase the volume of synthetic data generated significantly. In addition, we propose a method for generating anomalous situations, explained in more detail in Section \ref{subsec:anomalies}. This allows us to increase the heterogeneity of the synthetic data without additional effort or expert knowledge. Scenario prediction approaches solely aim to generate different realistic scenarios and examine the uncertainty between these scenarios to support predictive models. Moreover, the state-of-the-art proposals for GAN-based scenario generation do not take advantage of the multivariate data handling with complex nonlinear relationships that GANs allow. In contrast, in this work we generate scenarios to be used as a reliable source of synthetic data, and we make use of multivariate data with complex relationships and different natures (e.g., categorical). Table \ref{tab:soa-gan} summarizes the state-of-the-art limitations and the contributions of our proposal. To demonstrate the usefulness of the presented work, we will address a case study of real sensor data from an operating DC.

\section{Methodology}
\label{sec:method}

\subsection{Implemented  GAN Training Improvements}
\label{subsec:gan-train}

In the following, we describe the GAN training improvements found in the literature that have been implemented for this research. However, it should be noted that many of these enhancements have been proposed for image generation. Therefore, further research is needed to investigate their usefulness for time series generation. To further this purpose, some improvements mentioned above will be analyzed in Section \ref{sec:experiments}, in the hyperparameter tuning phase.

A GAN training phase consists of a minimax game between two neural networks, the Generator and the Discriminator (Equation \ref{eq:minimax-gan}). In this and the following equations, the Generator network is expressed as $G(z)$ and its parameters as $\theta_{g}$. The Discriminator network is expressed as $D(x)$ and its parameters as $\theta_{d}$. The sampling distribution of the data is expressed as $p_{\text {data}}$, and that of the Gaussian noise as $p(z)$. This process can be translated into a more general objective of the whole GAN architecture to make the generated data distribution look similar to the real data distribution. In practice, training a GAN is a complex and unstable process. During the last few years, many proposals have been made to solve some of its most common training issues. 

\begin{equation}
\min _{\theta_{g}} \max _{\theta_{d}}\left[\mathbb{E}_{x \sim p_{\text {data}}} \log D_{\theta_{d}}(x)+\mathbb{E}_{z \sim p(z)} \log \left(1-D_{\theta_{d}}\left(G_{\theta_{g}}(z)\right)\right)\right]
\label{eq:minimax-gan}
\end{equation}

Most algorithm stability problems during training are due to the Binary Cross Entropy (BCE) loss function. To solve this challenge, Arjovsky et al. \cite{arjovsky2017wasserstein} presented the Wasserstein GAN (WGAN), a novel GAN architecture implementing the Wasserstein loss function, which attempts to approximate Earth Mover's Distance (EMD). Since the Wasserstein function can take any real value (unlike the BCE function, which is limited between 0 and 1), the Discriminator is renamed the Critic. The WGAN solves many stability and fidelity problems. However, it requires a particular condition to approximate the EMD correctly: the loss function must be 1-Lipschitz continuous (i.e., the function's gradient norm must be less than or equal to 1 at all points). 

To ensure that 1-Lipschitz continuity is satisfied, WGAN authors suggested the Weight Clipping method \cite{arjovsky2017wasserstein}. They propose forcing the Critic network's weights to a fixed interval after updating through gradient descent algorithm. However, this solution needs intensive hyperparameter tuning to perform correctly. Ishaan Gulrajani et al. \cite{NIPS2017_892c3b1c} proposed a successful alternative: the Wasserstein GAN with Gradient Penalty (WGAN-GP). Consist in sampling some points by interpolating (using a random number $\epsilon$) between real and fake samples, and it is on the critics' prediction of these interpolated images that you want the 1-Lipschitz continuity condition to be met. Empirically, the Gradient Penalty method tends to produce higher quality results more stably. For its implementation, only a regularization factor needs to be added to the W-Loss cost function, weighted by a penalty coefficient $\lambda$ (Equation \ref{eq:wgan-gp}). In this equation, the interpolated data sample is expressed as $\hat{x}$, and the sampling distribution of the interpolated data as $p_{\hat{x}}$.

\begin{equation}
    \begin{array}{c}
    L_{d}=\underbrace{\mathbb{E}[D(x)]-\mathbb{E}[D(G(z))]}_{\text {Original critic loss }}+\underbrace{\lambda \cdot \mathbb{E}_{\hat{x} \sim p_{\hat{x}}}\left[\left(\| \nabla_{\hat{x}}\left(D(\hat{x}) \|_{2}-1\right)^{2}\right]\right.}_{\text {Gradient penalty }} \\
    \text { with } \hat{x}=\epsilon x+(1-\epsilon) G(z) \\
    \text { and } \epsilon \sim U[0,1]
    \end{array}
\label{eq:wgan-gp}
\end{equation}

Miyato et al. \cite{miyato2018spectral} proposed Spectral Normalization, another technique for improving GAN training's stability, which can be used in conjunction with WGAN-GP. It consists of normalizing the Critic weight matrices by their corresponding spectral norm, which also helps control the 1-Lipschitz continuity.

Heusel et al. \cite{NIPS2017_8a1d6947} proposed the Two Time-Scale Update Rule (TTUR). This method consists of using a higher learning rate in the Critic than in the Generator. The training is more stable and converges better towards Nash's equilibrium. The authors empirically demonstrate that this approach outperforms the original WGAN-GP proposal.

S. Chintala, a co-author of the prosperous DCGAN architecture \cite{yu_unsupervised_2017}, made a presentation at Neural Information Processing Systems (NIPS) 2016 \cite{david_lopez-paz_nips_2017} summarizing many tips for stable training of GANs. For instance: (i) normalize inputs between -1 and 1; (ii) sample noise from Gaussian distributions; (iii) build different batches for real and false data; (iv) use LeakyReLU activation function; (v) use Adam optimizer algorithm \cite{Kingma2015AdamAM}; (vi) use Embedding layers for discrete variables; (vii) use regularization methods such as Dropout \cite{srivastava_dropout_2014} and Batch Normalization \cite{Ioffe_batch_2015}.

All the above-mentioned GAN training improvements have been implemented for this research, and some (e.g., Dropout, TTUR) have been analyzed in the hyperparameter tuning phase to test their effectiveness on the results.

\subsection{On-Demand Anomalies}
\label{subsec:anomalies}
One of the significant advantages of using GAN over other algorithms in the literature to synthetically augment the data is that it gives us more control over the generation through the latent space. This feature can be particularly beneficial when working with unbalanced datasets. In the use case under study, we find that most of the samples change little from the previous step. To increase the heterogeneity of the data produced, we propose a method for generating on-demand anomalies, i.e., data where the time difference with the previous step is significantly high.

\begin{algorithm}
\caption{On-demand anomaly scenario generation algorithm}
\label{alg:anomaly}
\begin{algorithmic}
\item \textbf{Input:} Number of scenarios $S$, Lenght of each scenario prediction $L$, multidimensional time-series data $H$, categorical variable sensor Id $C$, Generator network $G$, anomaly occurrence step $\gamma$, standard deviation for anomalies $\sigma '$.

\State let Scenarios$[0,1,...,S-1]$ be a new array
\State let Prediction$[0,1,...,L-1]$ be a new array

\For {$s \in [0, S-1]$}

$X$ will be the time-series input to $G$:

$X \gets H$

    \For {$t \in [0, L-1]$}
    
    \If{$t = \gamma$}
    
    \hspace{0.8cm}$z \sim \mathcal{N}(0,\sigma ')$
    
    \Else
    
    \hspace{0.8cm}$z \sim \mathcal{N}(0,1)$
    
    \EndIf
    
    \hspace{0.5cm}Obtain the next step prediction: 
    
    \hspace{0.5cm}Prediction$[t] \gets G(X, C, z)$
    
    \hspace{0.5cm}Update $X$ by concatenating it with Prediction$[t]$: 
    
    \hspace{0.5cm}$X \gets X[1:]^\frown$Prediction$[t]$
    
    \EndFor
    
Allocate the obtained predictions as a new scenario:

Scenarios$[s] \gets$ Prediction$[\;]$

\EndFor

\end{algorithmic}
\end{algorithm}

The generation of new data using a GAN implies introducing a vector at the input of the Generator network. One of the improvements mentioned in the previous section recommends sampling these input vectors from a Gaussian distribution, parameterized by a mean of 0 and a standard deviation of 1. We propose increasing the standard deviation of the Gaussian noise distribution (during the generation phase) when the anomaly is desired, obtaining results that are far from the mean (i.e., unusual cases), leaving the standard deviation at 1 in the remaining time steps. Algorithm \ref{alg:anomaly} summarizes this process. After the computational analysis of Algorithm \ref{alg:anomaly} and a series of empirical tests, we can conclude that the computational complexity of the proposed method is $O(k\cdot n)$, where $k$ is a constant dependent on the number of scenarios generated and their length, and $n$ the input length.

Consequently, no expert knowledge is required to generate the anomalies. This method can generate "outlier" data for the models, which, despite losing certain realism (i.e., similarity to the real data), helps the model gain more stability. In Computer Vision tasks, data augmentation techniques that also make images "lose realism" (e.g., mixing, random erasing) have demonstrated their usefulness in numerous research studies \cite{summers_improved_2019} \cite{shorten_survey_2019}. However, its usefulness in time-series tasks has not yet been studied in depth. Our solution does not require additional efforts, and the physical integrity of the electronic equipment is not jeopardized by exposing them to extreme situations.

\subsection{Evaluation Metrics}
Validating the quality of the generated scenarios is more challenging than measuring the performance of point forecasts. On the one hand, scenarios must be realistic enough to reflect the structural and variable interdependencies of predicting values at different prediction horizons. On the other hand, they must have enough variability to provide valuable additional information when used as synthetic data. 

The most commonly used approach in the literature of synthetic time-series data generation is to compare the autocorrelation of the generated data with that of the real data. However, we consider this method to be limited for assessing the usefulness of synthetic data and is not suitable for multivariate data because it does not analyze the relationships between variables. This paper proposes to use two different metrics to verify the similarity of the generated multidimensional data: Kullback-Leibler (KL) Divergence and Mean Squared Error (MSE). 

\begin{equation}
    D_{\mathrm{KL}}(P \| Q)=\sum_{x \in \mathcal{X}} P(x) \log \left(\frac{P(x)}{Q(x)}\right)
    \label{eq:kl}
\end{equation}

The KL Divergence, also known as relative entropy, is an asymmetric measure of the difference between a given probability distribution $P$ and a reference probability distribution $Q$ (Equation \ref{eq:kl}). This metric gives us an overall assessment of the multidimensional distribution of the generated data compared to the real data, thus ensuring that the relationships between the variables are maintained. A KL divergence of 0 indicates that the two distributions are identical. In the context of coding theory, this metric can express the extra amount of bits (using $\log_2$) needed to code samples of a distribution $P$ using a code optimized for the compared distribution $Q$. In the Bayesian inference field, this divergence metric can express a measure of the information gained by revising one's beliefs from the prior probability distribution $Q$ to the posterior probability distribution $P$. A great advantage of this metric is that it can be used with multivariate data, using n-dimensional probability distributions.

\begin{equation}
    \mathrm{MSE}=\frac{1}{n} \sum_{i=1}^{n}\left(Y_{i}-\hat{Y}_{i}\right)^{2}
    \label{eq:mse}
\end{equation}

The MSE of an estimator measures the squares' average of the committed errors (Equation \ref{eq:mse}). Unlike the mean errors in absolute values, this metric gives greater weight to large errors. The MSE evaluates the punctual difference between the generated and actual time series, thus ensuring the temporal coherence of each variable. The lower the MSE score, the better the forecast.

Combining these two metrics aims to achieve realistic scenarios in the synthetic data generated, both in terms of multidimensional probability distribution and punctual prediction of each variable.

\section{Case Study}
\label{sec:case-study}
This section describes the use case employed to validate our proposal and the complete architecture implemented. In this research, we have used real sensor data collected from an operating DC of the company \textit{Adam Data Centers}\footnote{\textit{Adam Data Centers} [\url{https://adam.es/data-center/}]}. The data was collected during 2019 at \textit{Adam Data Centers}' facility in Navalcarnero (Madrid, Spain). The sensors were designed and developed by the startup company \textit{TycheTools}\footnote{\textit{TytheTools} [\url{https://www.tychetools.com}]}.

\begin{figure}[htb]
  \center
  \caption{Plot of the data collected by one sensor. Samples are collected every 10 minutes.}
  \includegraphics[width=0.75\textwidth,keepaspectratio]{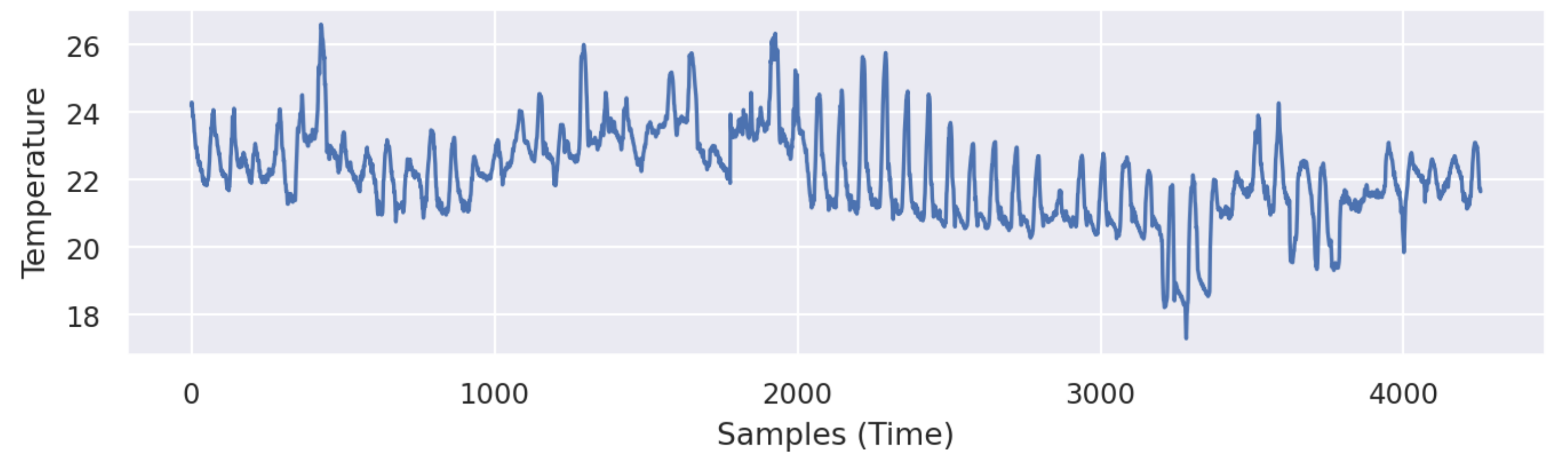}
  \includegraphics[width=0.75\textwidth,keepaspectratio]{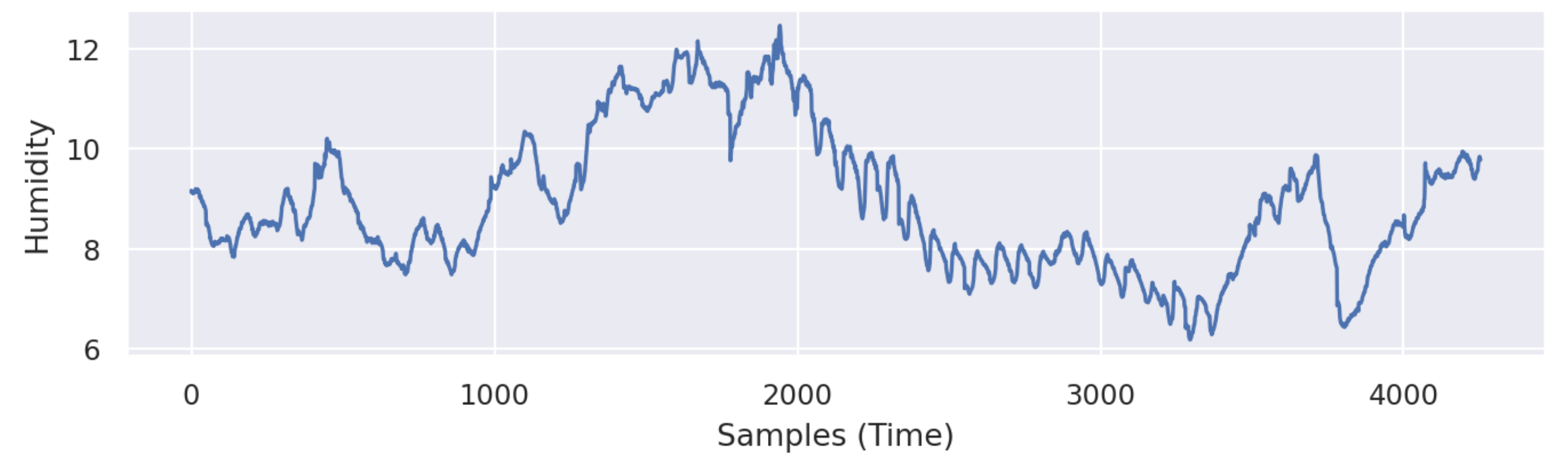}
  \label{fig:plot_sensor}
\end{figure}

DC facilities are primarily based on air-based cooling solutions, following the Hot Aisle / Cold Aisle strategy \cite{noauthor_hot}. It consists of lining up the racks (i.e., computing units) to face cold air intakes and hot air exhausts alternately, forming cold and hot aisles. The air conditioning units cool down the heated air from the hot aisles and then reintroduce it to the cold aisles using the ventilation ducts on the room’s floor. The sensors were placed equally in the cold and hot aisles in the \textit{Adam Data Centers}' facility. To model a DC facility’s behavior, it is crucial to monitor as many system variables as possible. However, temperature and relative humidity are the most relevant and straightforward variables to capture. Besides studying the cooling system dynamics, there is a clear correlation between these variables and the total power consumed \cite{rahmani2018complete} \cite{Tsilingiris2008ThermophysicalAT}. Considering them sufficiently relevant, only relative humidity and temperature were captured. Nevertheless, following our approach, the number of variables can be easily expanded.

\begin{figure*}[htb]
  \center
  \caption{On the left: Rolling segmentation on dataset. The green bars represent the entire dataset, the blue bars represent the model inputs, and the orange bars represent the model outputs. On the right: Kernel Density Estimation plot confronting Relative Humidity and Temperature data from all sensors.}
  \includegraphics[width=\textwidth,keepaspectratio]{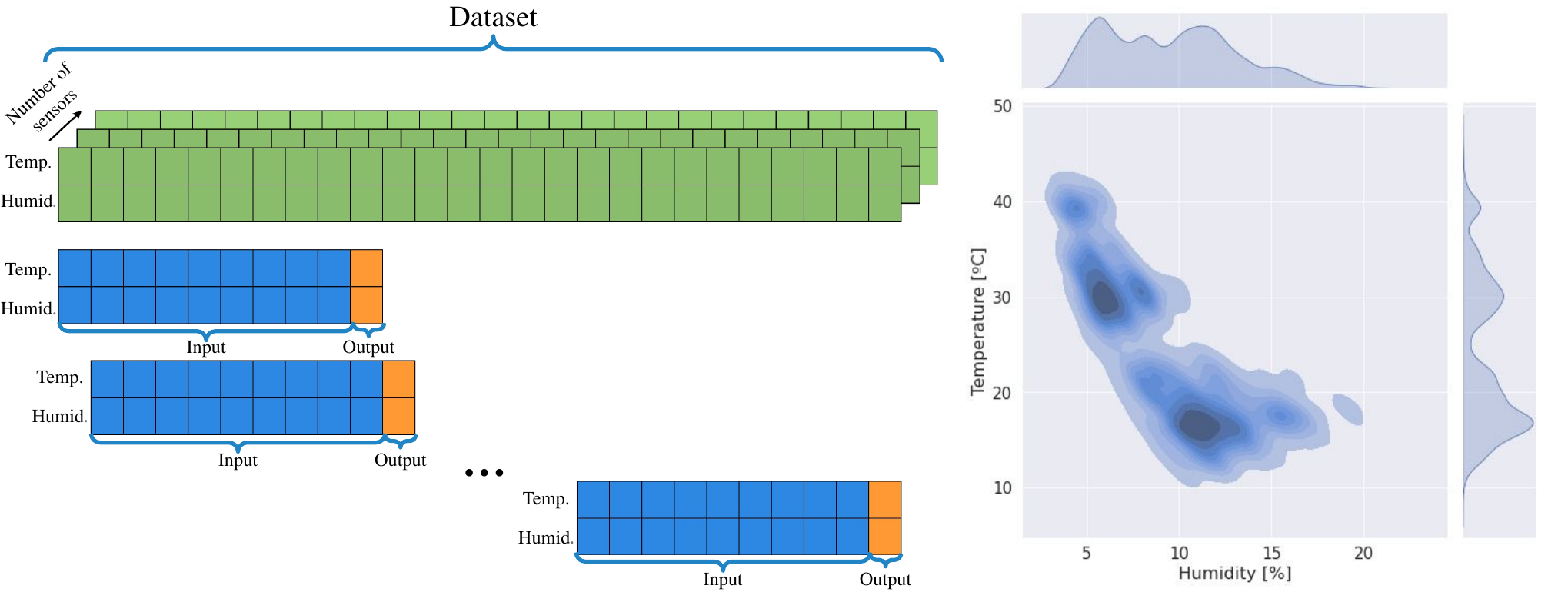}
  \label{fig:plot_KDE}
\end{figure*}

We have a total of 35 sensors available. The measurements are taken every 10 minutes and sent to the gateway that inserts them into the database. In Figure \ref{fig:plot_sensor}, we can observe the data collected by one of the sensors. We can observe a certain negative correlation by comparing both variables (humidity and temperature) from all sensors and computing the Kernel Density Estimation (Figure \ref{fig:plot_KDE}). These two variables show a Pearson correlation coefficient of -0.784. When the humidity is low, higher temperatures are registered, and vice-versa. This phenomenon occurs because water vapor concentration impacts the air's thermal conductivity, affecting the power consumption of cooling systems \cite{Tsilingiris2008ThermophysicalAT}.

\begin{figure}[htb]
  \center
  \caption{Complete GAN architecture.}
  \includegraphics[width=0.8\textwidth,keepaspectratio]{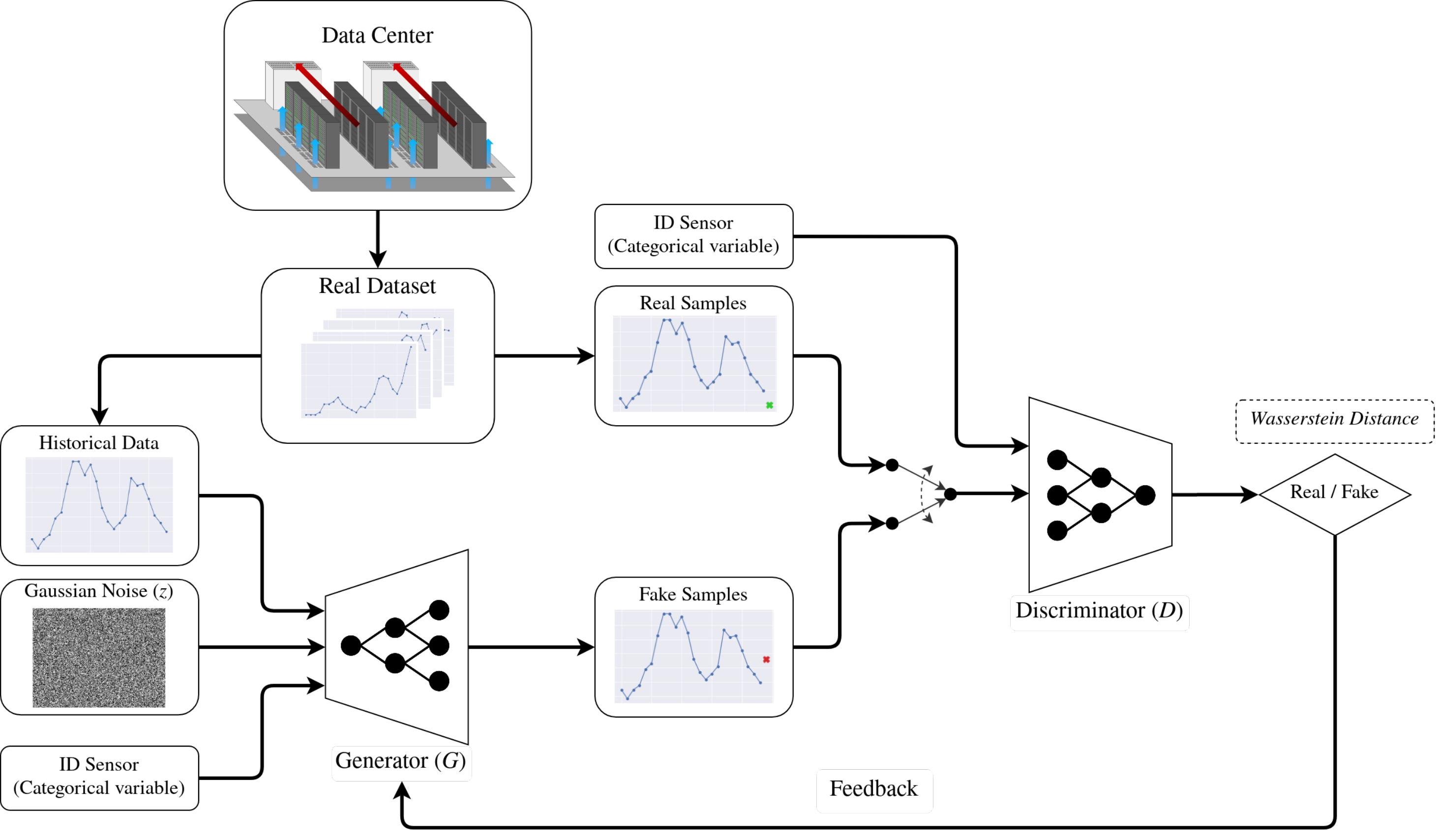}
  \label{fig:gan_arch}
\end{figure}

We segmented the dataset with a fixed-size rolling window (Figure \ref{fig:plot_KDE}). Figure \ref{fig:gan_arch} shows the complete GAN architecture, specifying each input and output. The architecture is designed to introduce one sensor at a time, which will give us high flexibility in the data generation phase. Although, due to this decision, the model may have more difficulties finding the correlations between the different sensors.

The Generator network has three inputs: (i) the conditional categorical input: a variable indicating the sensor ID, treated by an embedding layer; (ii) the conditional time-series input: a multidimensional historical data from the sensor, in our case, humidity and temperature, treat by LSTM neurons; (iii) Random noise input: vectors sampled from a multidimensional Gaussian distribution (i.e., latent space), treat by Fully Connected neurons. With this approach, the Generator will produce synthetic data based on the random noise input, conditioned to the particular sensor being handled and the temperature and humidity data it has measured in the last hours. Therefore, the Generator network must reproduce the patterns and relationships between the conditioning variables. The Critic network has two inputs: (i) the conditional categorical input: a variable indicating the sensor ID, treated by an embedding layer of the same dimension as in the Generator network; (ii) the data input: multidimensional time-series data, alternating between real and fake data.

The Generator network produces a multidimensional output, attempting to predict the next step of the introduced time series correctly. Therefore, if we wish to obtain longer-term predictions, we must reintroduce the previously produced output to the input data and repeat it to the desired point. Once the training phase is completed, we will analyze the results by exploring the latent space. We will also illustrate the generation of on-demand anomalous situations to increase the data's variability.

\section{Experiments and Results}
\label{sec:experiments}
This section describes the performed experiments and discusses the quality of the obtained results. The code developed for this research was developed in the \textit{Google Colab} platform, using \textit{Python} 3.6, and the \textit{TensorFlow} DL framework. The decision to develop the project using the \textit{Google Colab} platform is primarily due to its free-charge Graphics Processing Unit (GPU) executions to train models. Besides, this platform encourages the projects’ reproducibility and compliance with open science’ best practices by offering ubiquitous programming virtual environments. The complete code is openly available on \textit{GitHub} \cite{perez_gan_scenario_forecasting_2021}.

As explained in Section \ref{subsec:gan-train}, GAN training is a complex and delicate process. Training stability depends on many hyperparameters that must be set by hand and have complex interrelationships. In image generation, the research community has made a great effort to find the hyperparameters that produce better results and more stable training. Unfortunately, the use of GANs for time-series data has been scarcely explored, and no in-depth studies have explored the best hyperparameters for this field. Therefore, in this work we have employed many recommendations from the GANs’ state of the art, combined with an empirical heuristic search to set some reasonably good initial hyperparameters reflected in Table \ref{tab:hyper-init}.

\begin{table}[]
    \center
    \tiny
    \caption{Initial GAN training hyperparameters}
    \begin{tabular}{cc}
    \toprule
    Hyperparameters & Values \\ \midrule
    Feature Scaling & Min-Max Scaler {[}-1, 1{]} \\
    Training Loss Function & \makecell{Wasserstein Loss \\ with Gradient Penalty} \\
    Batch Norm in Generator network & \cmark \\
    Spectral Normalization & \cmark \\
    Gaussian Noise Dimension & 8 \\
    Embedding Layer Dimension & 8 \\
    Networks Size Ratio (Critic / Generator) & $\sim$4.5 \\
    Batch Size & 64 \\ \bottomrule
    \end{tabular}
    \label{tab:hyper-init}
\end{table}

\begin{table*}

\centering
\scriptsize
\caption{GAN training hyperparameter tuning process. Experiments result using validation set (10\% of the dataset).}
\label{tab:exp-results}
\scalebox{0.65}{%
\begin{tabular}{cccccccc} 
\toprule
\multicolumn{5}{c}{\textbf{Hyperparameters}}                                                                                                                                                                                                                                                             & \multicolumn{3}{c}{\textbf{Results Metrics}}                                                                                                            \\ 
\midrule
\multirow{2}{*}{\textbf{Optimizer}} & \multirow{2}{*}{\begin{tabular}[c]{@{}c@{}}\textbf{Skip}\\\textbf{ Connection}\end{tabular}} & \multirow{2}{*}{\begin{tabular}[c]{@{}c@{}}\textbf{Output}\\\textbf{ Activation}\end{tabular}} & \multirow{2}{*}{\textbf{TTUR}} & \multirow{2}{*}{\textbf{Dropout}} & \multirow{2}{*}{\begin{tabular}[c]{@{}c@{}}\textbf{KL Divergence} \\ {[}bits]\end{tabular}} & \multicolumn{2}{c}{\textbf{MSE}}                          \\ 
\cline{7-8}
                                    &                                                                                              &                                                                                                &                                &                                   &                                                                                             & \textbf{Temp. [$^\circ C^2$]} & \textbf{Humid. [$\%^2$]}  \\ 
\midrule
Adam                                & \xmark                                                                        & \textit{linear}                                                                                & \xmark          & \xmark             & 1.888                                                                                       & 1.943                         & 1.011                     \\
Adam                                & \xmark                                                                        & \textit{linear}                                                                                & \xmark          & \cmark             & 1.801                                                                                       & 2.359                         & 1.092                     \\
Adam                                & \xmark                                                                        & \textit{linear}                                                                                & \cmark          & \xmark             & 1.771                                                                                       & 2.206                         & 0.778                     \\
Adam                                & \xmark                                                                        & \textit{linear}                                                                                & \cmark          & \cmark             & 1.825                                                                                       & 2.468                         & 0.844                     \\
Adam                                & \xmark                                                                        & \textit{tanh}                                                                                  & \cmark          & \xmark             & 2.431                                                                                       & 2.203                         & 0.987                     \\
Adam                                & \cmark                                                                        & \textit{linear}                                                                                & \cmark          & \xmark             & 2.358                                                                                       & 2.244                         & 0.721                     \\
AdaBelief                           & \xmark                                                                        & \textit{linear}                                                                                & \xmark          & \xmark             & 2.375                                                                                       & 3.134                         & 1.083                     \\
AdaBelief                           & \xmark                                                                        & \textit{linear}                                                                                & \xmark          & \cmark             & 2.013                                                                                       & 1.473                         & 0.662                     \\
AdaBelief                           & \xmark                                                                        & \textit{linear}                                                                                & \cmark          & \xmark             & 1.707                                                                                       & 1.359                         & 0.583                     \\
AdaBelief                           & \xmark                                                                        & \textit{linear}                                                                                & \cmark          & \cmark             & {\color{blue} \textbf{1.432}}                                                                              & {\color{blue}\textbf{0.977}}                & {\color{blue}\textbf{0.438}}            \\
AdaBelief                           & \xmark                                                                        & \textit{tanh}                                                                                  & \cmark          & \cmark             & 1.872                                                                                       & 1.317                         & 0.419                     \\
AdaBelief                           & \cmark                                                                        & \textit{linear}                                                                                & \cmark          & \cmark             & 2.018                                                                                       & 1.556                         & 1.187                     \\
\bottomrule
\end{tabular}
}
\end{table*}

After the first hyperparameter selection process, and based on state-of-the-art examples and some empirical tests, we decided to use LSTM neurons in the Generator and 1D Convolution neurons in the Critic. The number of trainable parameters in the Critic network is approximately 735,000, and in the Generator network, around 165,000.

Once the architectures are fixed, we adjust the following hyperparameters: (i) Optimizer algorithm: Adam \cite{Kingma2015AdamAM} or AdaBelief \cite{zhuang2020adabelief}; (ii) Skip-Connection: An additional connection in the Critic network inspired by the ResNet architecture \cite{he2016deep} architectures; (iii) Activation function at the Generator output: \textit{Linear} or \textit{tanh}; (iv) TTUR \cite{NIPS2017_8a1d6947}; (v) Dropout \cite{srivastava_dropout_2014}. 

Table \ref{tab:exp-results} illustrates the results of this hyperparameter adjustment. The metrics shown are performed by comparing a random scenario generation of 24 time-steps ahead (4 hours) with the ground truth from a randomly selected validation set (10\% of the dataset). 

\begin{figure*}[htb!]
  \center
  \caption{On the left: Diagram showing how the "Real" and "Generated" sets of data were obtained. The Generated data consist of 24-steps random scenarios from a validation set of data (10\% of the dataset). On the right: Kernel Density Estimation plot confronting Real and Generated data distributions.}
  \includegraphics[width=\textwidth,keepaspectratio]{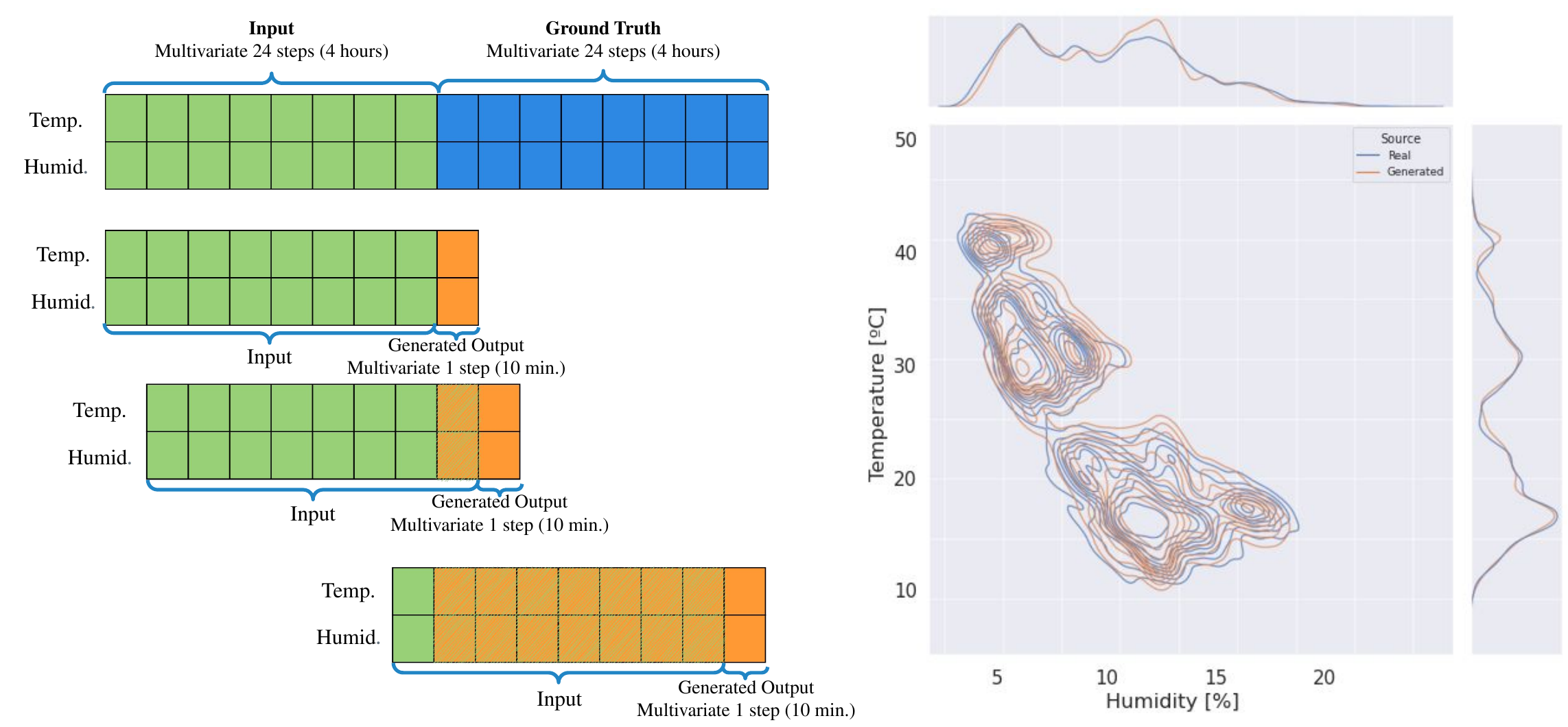}
  \label{fig:results_kde}
\end{figure*}

We have achieved the following results in a validation set (10\% of the dataset): KL divergence of 1.432 bits and a Root Mean Squared Error (RMSE) accuracy error of 0.988°C for temperature and 0.661\% for humidity. Figure \ref{fig:results_kde}, using the Kernel Density Estimation, illustrates the comparison of the distributions from the scenarios predicted of 24 time-steps (4 hours) ahead and the ground truth from a randomly selected validation set. Hence, the predicted scenarios appear consistent with the available real data. 

To demonstrate the quality of the results generated by our model, we have conducted experiments in a randomly selected test set (15\% of the dataset), achieving a KL divergence of $1.363\pm0.071$ bits and an RMSE accuracy error of $0.995\pm0.08$°C for temperature and $0.661\pm0.03$\% for humidity, very similar results to those obtained in the validation set (Table \ref{tab:exp-results}).

It should be noted that a comparison of these results with the methods discussed in the literature would be meaningless since, as we have explained in Section \ref{subsec:gan_soa}, the reviewed approaches focus only on similarity metrics (e.g., autocorrelation). However, our goal is to achieve both similarity and variability in the scenarios to obtain as much synthetic data as possible, taking advantage of the possibilities offered by GANs through the exploration of their latent space.

The obtained results are indicators that the synthetic data we have generated have high similarity (a standard measure of realism) to the actual data. However, we are not guaranteed that these generated data have significant variability or help train predictive or optimization models. To this end, in the following subsections, we will visually explore examples of the scenario and on-demand anomaly generation and evaluate their utility by training a predictive model.

\subsection{Scenario Generation}
\label{subsec:random_scen}
During the GAN training phase, we introduced noise vectors sampled from a multidimensional Gaussian space, known as latent space. This procedure implies that when creating a new scenario, the Generator's output will change depending on the noise vector introduced at the input. Therefore, by randomly sampling vectors from latent space, and if the GAN has trained correctly, we get different realistic scenarios for the same given situation. 

\begin{figure}[htb]
  \center
  \caption{Scenario Generation example of 24-steps (4 hours) ahead. Example with low uncertainty. The scenarios were generated by concatenating (reintroducing) each 1-step prediction at the input of the Generator.}
  \includegraphics[width=0.7\textwidth,keepaspectratio]{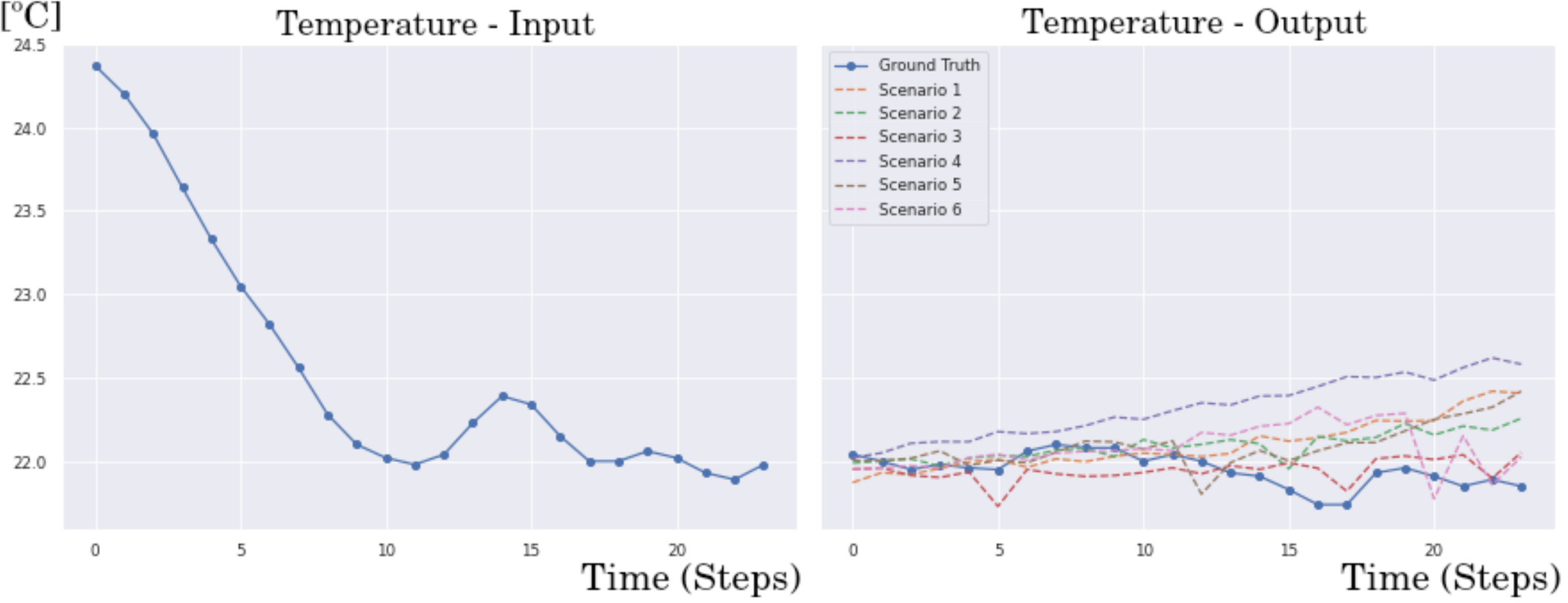}
  \includegraphics[width=0.7\textwidth,keepaspectratio]{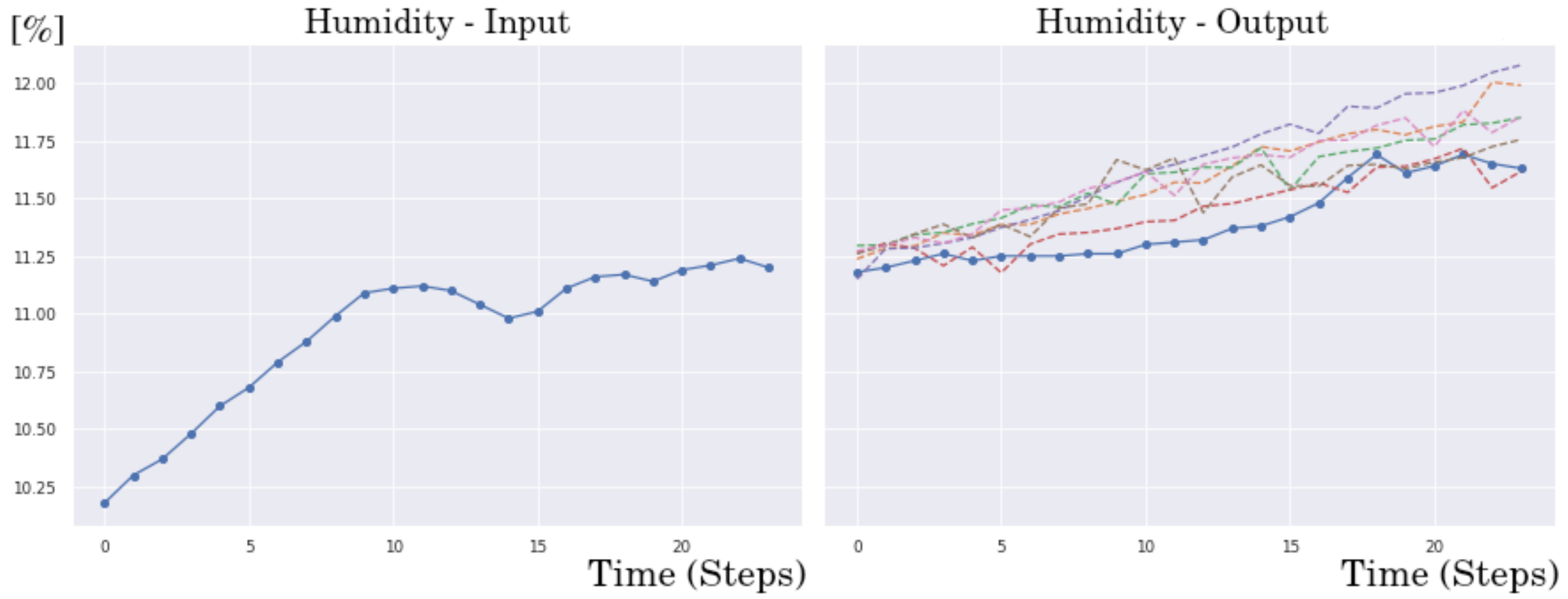}
  \label{fig:scen_1}
\end{figure}

\begin{figure}[htb]
  \center
  \caption{Scenario Generation example of 24-steps (4 hours) ahead. Example with increasing uncertainty. The scenarios were generated by concatenating (reintroducing) each 1-step prediction at the input of the Generator.}
  \includegraphics[width=0.7\textwidth,keepaspectratio]{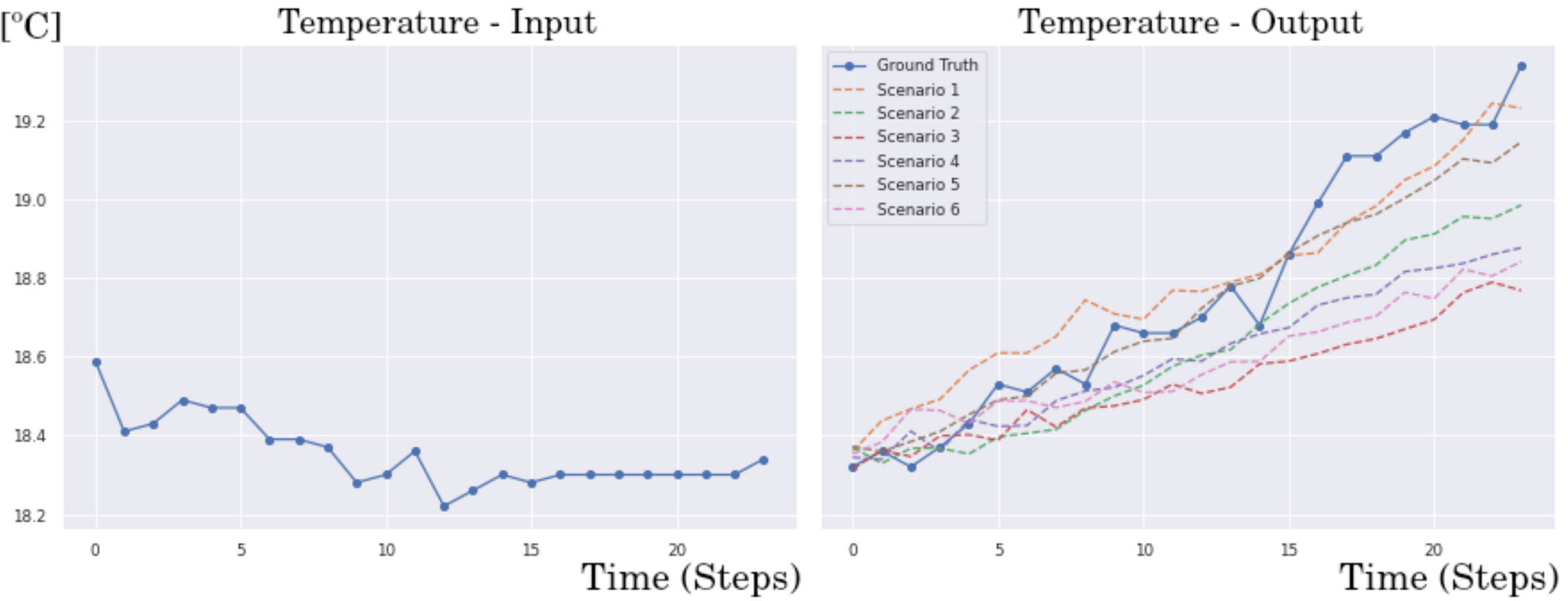}
  \includegraphics[width=0.7\textwidth,keepaspectratio]{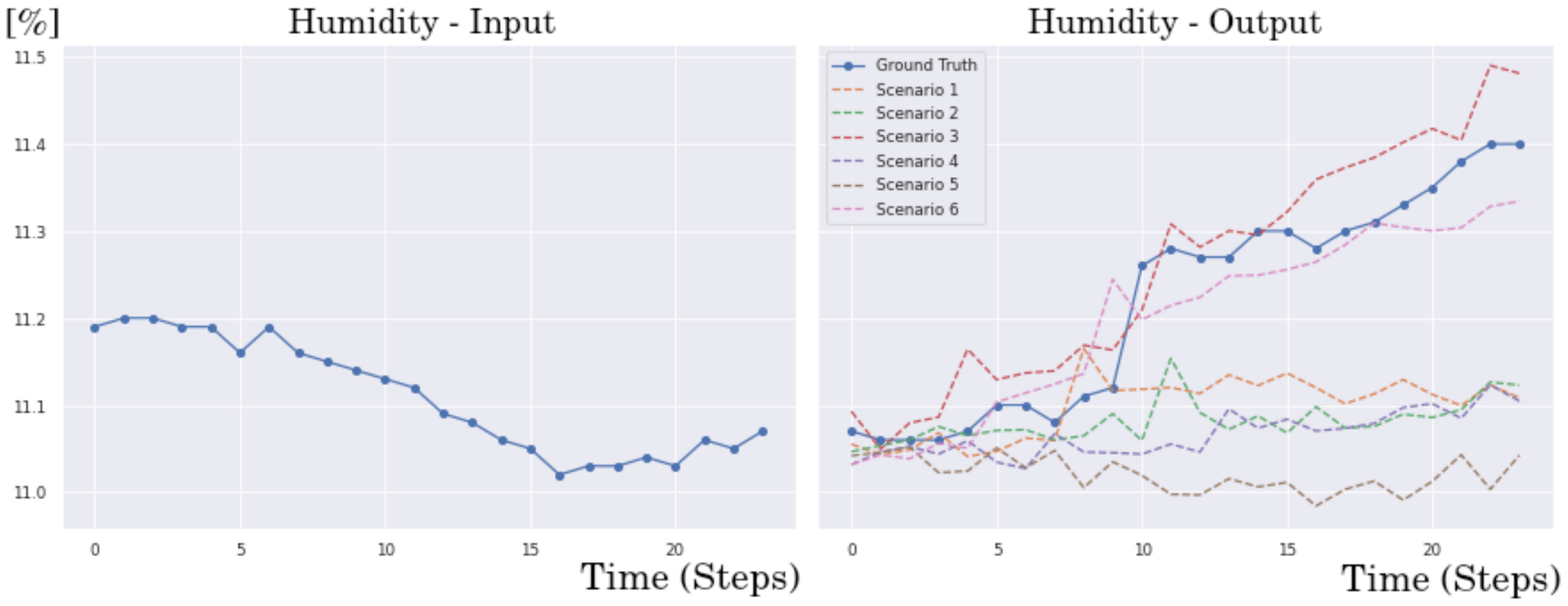}
  \label{fig:scen_2}
\end{figure}

In the following, we find some examples of scenarios generated, which help us visualize the results' quality and the achieved variability. Specifically, six different scenarios have been generated in each example, with a duration of 24 time-steps concatenated predictions (4 hours ahead). In Figure \ref{fig:scen_1}, we present a situation where the scenarios exhibit low uncertainty, indicating that the Generator has higher confidence in the results. Figure \ref{fig:scen_2} shows scenarios where the uncertainty is relatively low in early prediction steps, but it increases as we advance in the number of steps ahead. We can observe how the correlation between temperature and humidity variables remains consistent in all the generated scenarios. 

\begin{figure*}[htb!]
  \center
  \caption{On the left: Heat maps showing the data collected by the 35 available sensors during 24 time steps (4 hours). Each row represents a different sensor. On the right: Heat map showing a scenario generated in the 35 available sensors. 24-step prediction (4 hours ahead). Each row represents a different sensor.}
  \includegraphics[width=0.88\textwidth,keepaspectratio]{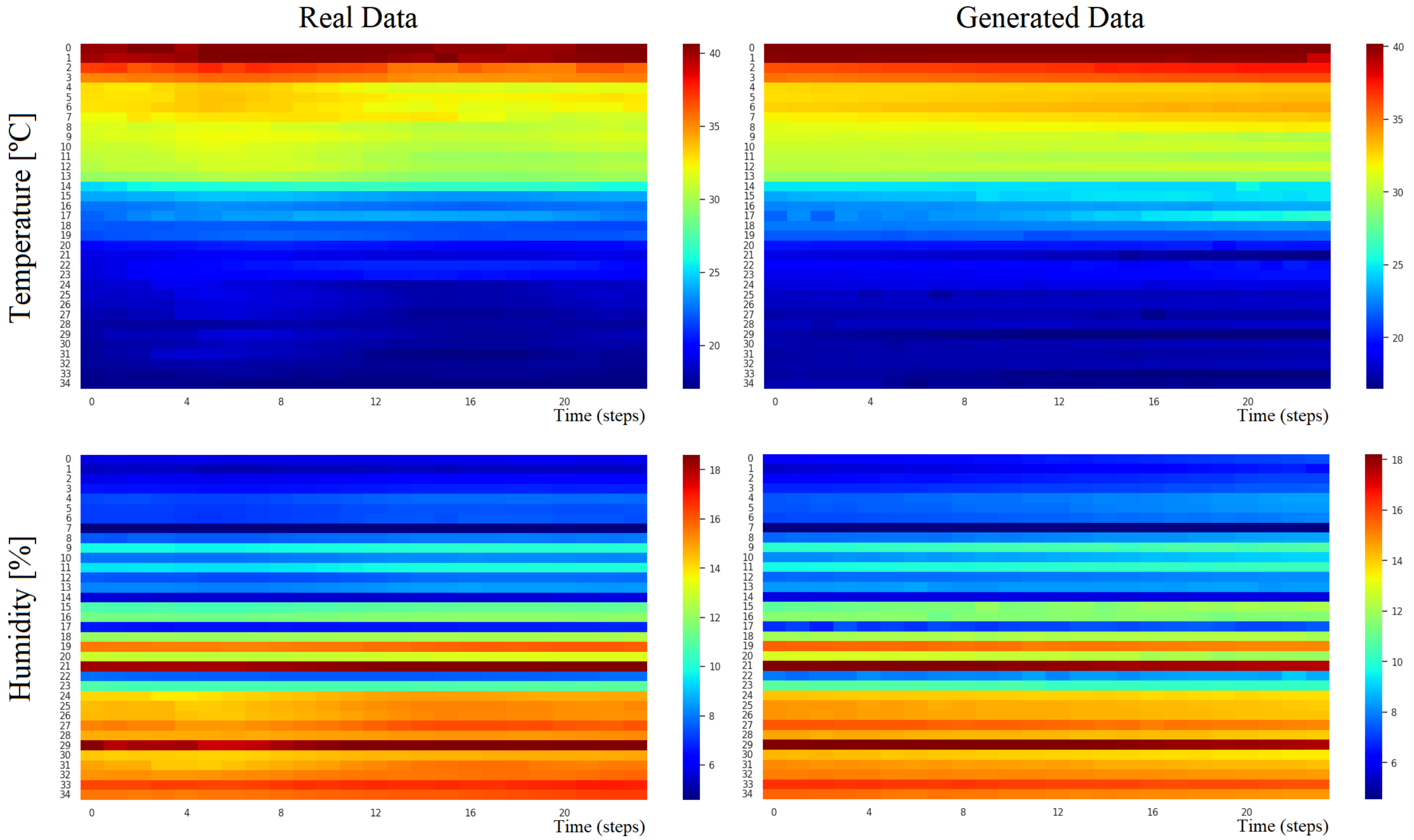}
  \label{fig:heatmap}
\end{figure*}

To provide an overview of a scenario generated on all sensors, in Figure \ref{fig:heatmap}, we find heat maps where each row represents a different sensor. On the right side of the Figure \ref{fig:heatmap}, we show the data generated with a duration of 24 time-steps predictions (4 hours ahead). On the left are the actual data collected. The rows (representing each sensor) have been sorted from highest to the lowest initial temperature.

\begin{figure}[htb!]
  \center
  \caption{Hot aisle/cold aisle scheme used in DC cooling systems.}
  \includegraphics[width=0.4\textwidth,keepaspectratio]{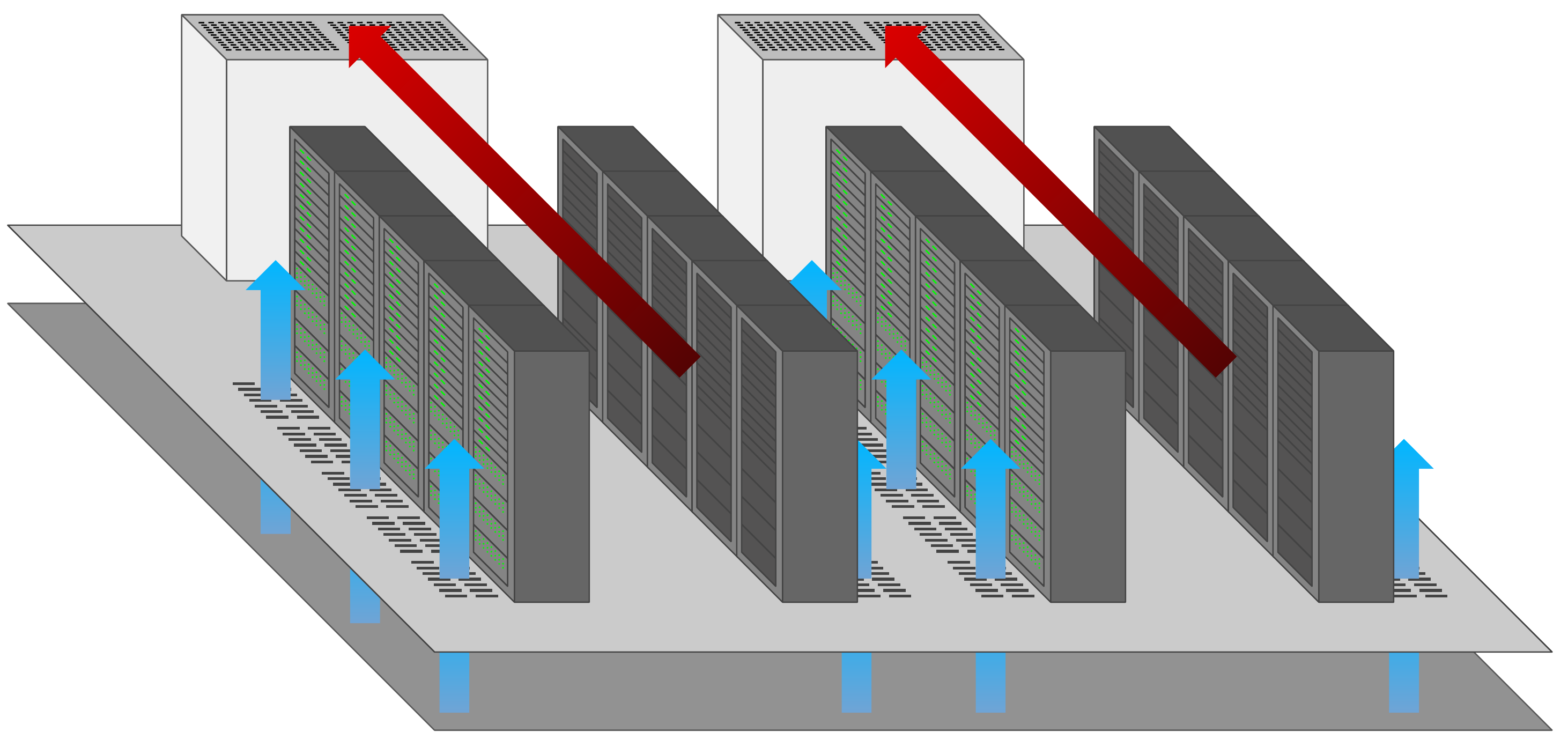}
  \label{fig:DC_raised_floor}
\end{figure}

The significant differences in temperature (and humidity) between the sensors result from their position in the DC. The higher temperature sensors are located in the hot aisles of the DC (server outlets), and the lower temperature sensors are located in the cold aisles (server inlets). The cold air enters the servers and flushes the heat generated by the computation to the outside. Moreover, as the workload is not homogeneously distributed, the temperature of a server's outlet also varies depending on the computation performed by the adjacent servers. On the other hand, the temperature of the inlets is not homogeneous either, as it depends mainly on the height at which the server is located. The cold air is driven through the raised floor and ascends to be absorbed by the server's front. The air's temperature reaching the servers at the top of a rack is usually higher than at the bottom. Figure \ref{fig:DC_raised_floor} shows a diagram of this scheme widely used in DC cooling systems. The heat maps in Figure \ref{fig:heatmap} show how this heterogeneity in the data is consistently maintained by the generated data, thus indicating that our approach allows the generation of realistic data coherent with the particularities of the use case under study.

\begin{table}[htb!]
\center
\caption{Utility evaluation, training a prediction model. Train on Synthetic, Test on Real (TSTR); Train on Synthetic and Real, Test on Real (TSRTR); Train on Real, Test on Real (TRTR)}
\begin{tabular}{cccc}

\toprule
                & TSTR          & TSRTR                                          & TRTR   \\ \hline
MSE Temperature & $0.106\pm 0.013$ & {\color{blue} \textbf{$0.0525\pm 0.004$}} & $0.0738$ \\
MSE Humidity    & $0.018\pm 0.005$ & {\color{blue} \textbf{$0.0087\pm 0.001$}} & $0.0094$ \\
\bottomrule
\end{tabular}
\label{tab:TRTSresults}
\end{table}

To empirically test the effectiveness of the data generated for training prediction or optimization models, we have followed the approach proposed by C. Esteban et al. \cite{esteban_real-valued_2017} of comparing the metrics achieved in Test on Synthetic, Test on Real (TSTR) with Train on Real, Test on Real (TRTR). This approach has also been used in other related work in the state of the art \cite{Fekri_Generating_2020}. Additionally, we compared these metrics with Train on Synthetic and Real, Test on Real (TSRTS) to test the effectiveness of combining both synthetic and real datasets during the training phase. 

We have developed a simple model based on feedforward neural networks to perform these tests. This model has 2 hidden layers with 256 and 32 neurons, respectively, and an output layer with 2 neurons (one for each variable to be predicted). To facilitate the evaluation, the objective of this model will be to predict the next step of the multivariate time series. Specifically, in each performed experiment, we have generated 15,000 data samples of 24 time steps, and on each sample, we have generated 10 different scenarios. Making a total of 150,000 data samples of 24 time steps, which is a larger volume than the original dataset. Table \ref{tab:TRTSresults} shows the results of the conducted tests. Since the synthetic data generation phase involves a certain degree of randomness, the results using synthetic data in training show the mean of 5 experiments and the confidence interval. We can observe that training with synthetic data only (TSTR) provides similar results to training solely with real data (TRTR). Nevertheless, both are satisfactorily outperformed by simultaneously training with synthetic and real data (TSRTR). These results are consistent with the conjecture that in the field of DL, a larger amount of useful data improves the performance of the models.

\subsection{On-Demand Anomaly Generation}
As explained in Section \ref{subsec:random_scen}, we have introduced randomly sampled vectors from a multidimensional Gaussian space during each training step, employing a distribution with a 0 mean and standard deviation of 1. Our proposal for generating on-demand anomalies (Section \ref{subsec:anomalies}) consists of increasing the standard deviation of the Gaussian noise distribution (during the generation phase) when the anomaly is desired, obtaining results distant from the mean (i.e., unusual cases).

\begin{figure}[htb]
  \center
  \caption{On-demand anomaly generation on the 10th step of the prediction. 24-steps prediction ahead (4 hours).}
  \includegraphics[width=0.7\textwidth,keepaspectratio]{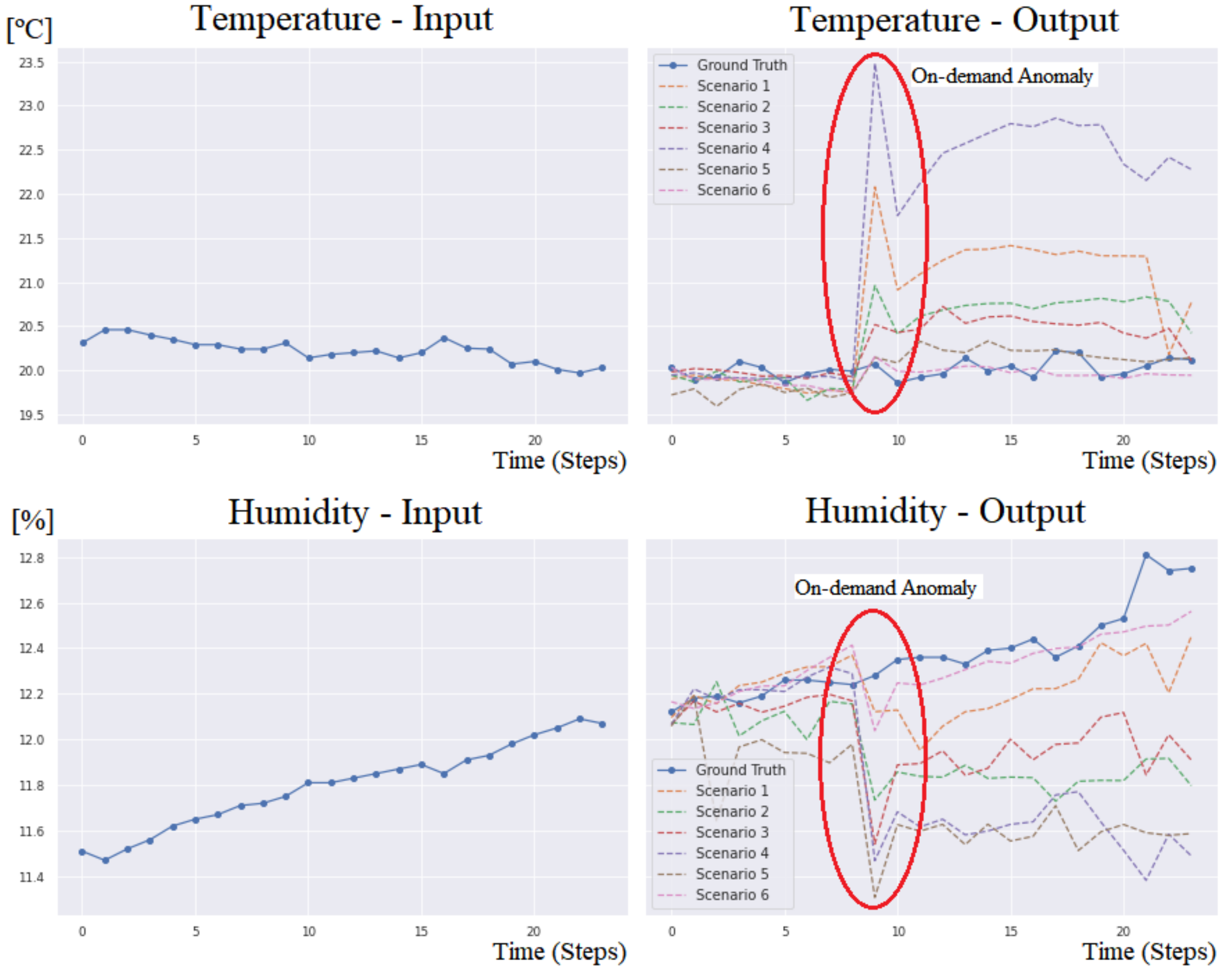}
  \label{fig:anomaly_scen}
\end{figure}

Figure \ref{fig:anomaly_scen} shows an example in which we have introduced an anomaly in the 10th prediction step by increasing the standard deviation from 1 to 8. We can appreciate that some scenarios produce the desired anomaly and that the following steps of the forecasting are consistent with this event. Not all scenarios produce anomalies because we are solely increasing the standard deviation of the Gaussian distribution. Thus, some sampled vectors will be in the same range (or close) as those used during the GAN training phase.

\begin{figure}[htb]
  \center
  \caption{Temperature and humidity derivative histograms indicating the difference between contiguous data points. The red dashed lines indicate $\pm5$ times the standard deviation ($\sigma$).}
  \includegraphics[width=0.48\textwidth,keepaspectratio]{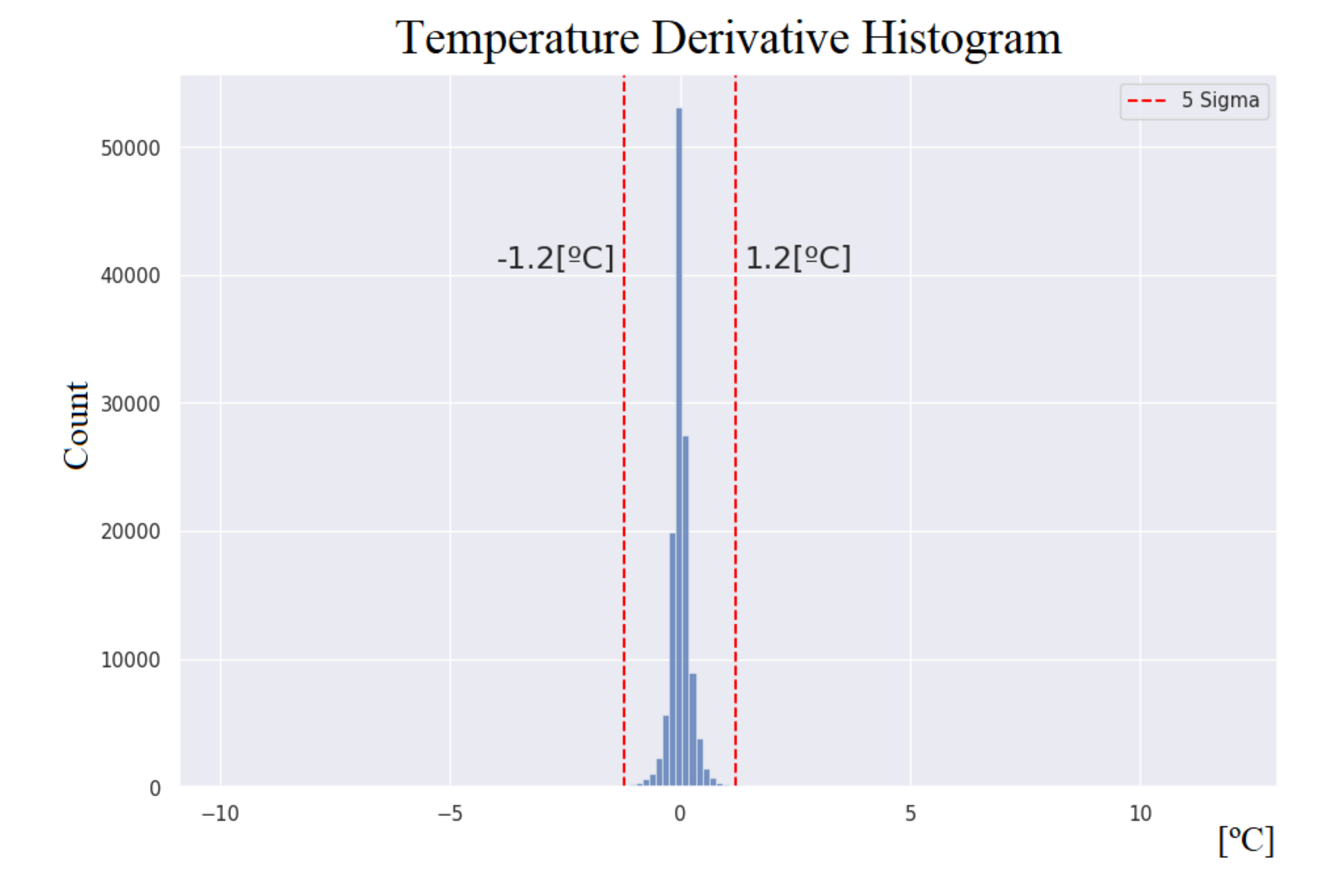}
  \includegraphics[width=0.47\textwidth,keepaspectratio]{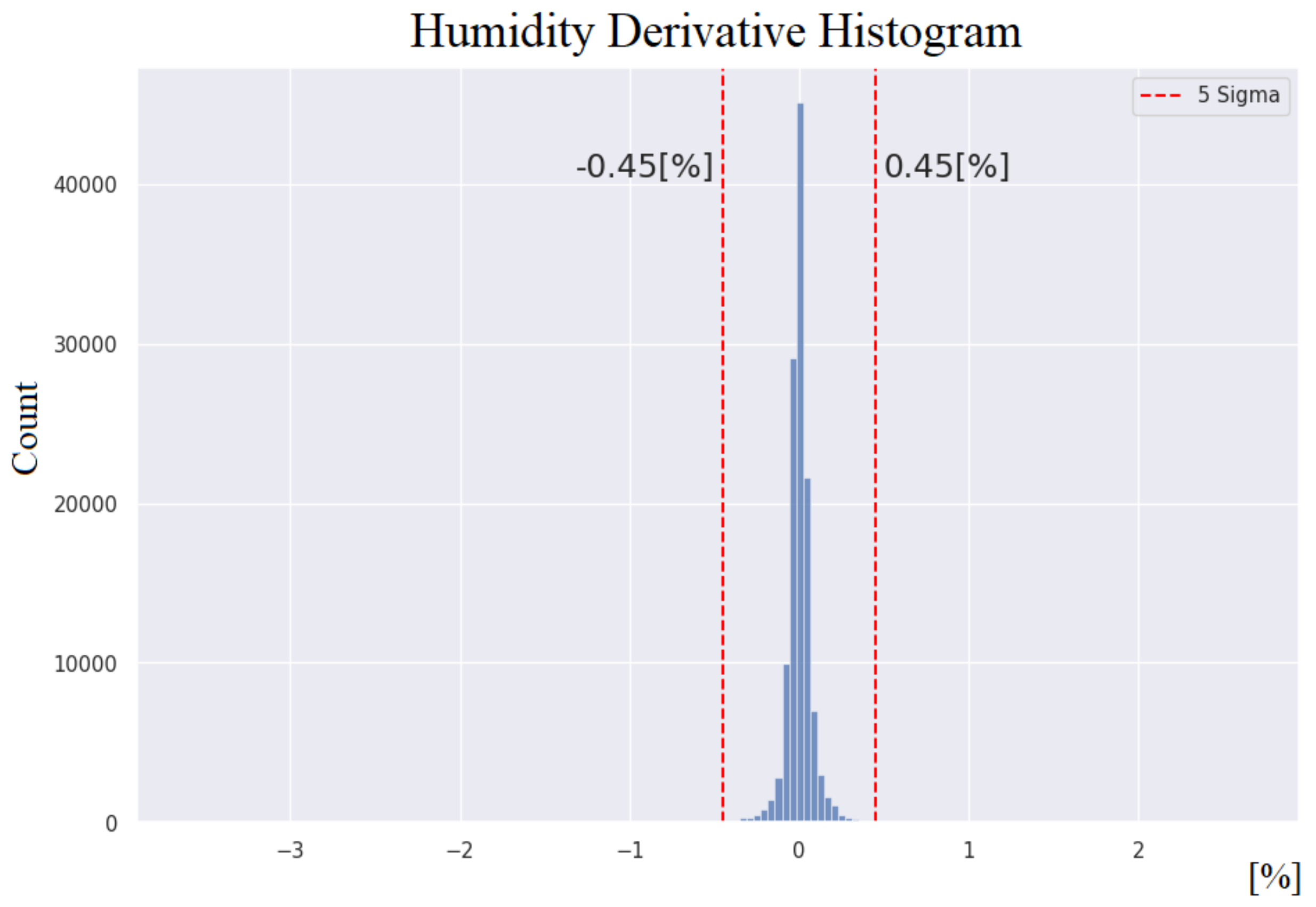}
  \label{fig:sigma}
\end{figure}

To precisely determine whether we have successfully generated an anomaly, we propose a threshold of $\pm5$ times the standard deviation ($\sigma$) of the derivative on each variable. This means that we observe if the difference between one data point and the next one is exceptionally anomalous (5$\sigma$). For temperature, this value is $\pm1.2$°C, and for humidity, it is $\pm0.45$\% (Figure \ref{fig:sigma}). In Figure \ref{fig:sigma}, we notice that for the vast majority of contiguous data points, the difference is close to 0. This makes sense because workloads in data centers do not usually undergo abrupt changes. These abrupt changes could be indicators of server failures or poor workload distribution in the DC. The maximum value measured in the temperature derivative is 11.9°C, and the minimum is -9.79°C. In the case of the humidity derivative, the maximum value is 2.64\% and the minimum -3.59\%.

\begin{table}[]
\center
\tiny
\caption{Evaluation of the generated anomalies, training a prediction model. Train on Synthetic, Test on Real (TSTR); Train on Synthetic and Real, Test on Real (TSRTR); Train on Real, Test on Real (TRTR)}

\begin{tabular}{cccccc}
\toprule
                                                          & \begin{tabular}[c]{@{}c@{}}TSTR\\ (5\% Anomalies)\end{tabular} & \begin{tabular}[c]{@{}c@{}}TSTR\\ (10\% Anomalies)\end{tabular} & \begin{tabular}[c]{@{}c@{}}TSRTR\\ (5\% Anomalies)\end{tabular} & \begin{tabular}[c]{@{}c@{}}TSRTR\\ (10\% Anomalies)\end{tabular} & TRTR   \\ \hline
\begin{tabular}[c]{@{}c@{}}MSE\\ Temperature\end{tabular} & $0.103\pm 0.008$                                                  & $0.0899\pm 0.004$                                                 & {\color{blue} \textbf{$0.0532\pm 0.004$}}                           & $0.0584\pm 0.005$                                                   & $0.0738$ \\
\begin{tabular}[c]{@{}c@{}}MSE\\ Humidity\end{tabular}    & $0.0168\pm 0.003$                                                 & $0.0139\pm 0.002$                                                 & {\color{blue} \textbf{$0.0078\pm 0.004$}}                           & $0.0086\pm 0.002$                                                   & $0.0094$ \\
\bottomrule
\end{tabular}
\label{tab:TSTR_Anomaly_Results}
\end{table}

As in Section \ref{subsec:random_scen}, we have evaluated the generated anomalies using the method proposed by C. Esteban et al. \cite{esteban_real-valued_2017}, in which the metrics of TSTR, TSRTR, and TRTR are compared in a predictive model. In this case, the anomalies are introduced in a certain percentage of the synthetic data with which the model is trained. That is, we do not train the model solely on anomalous situations. It is crucial to note that the objective of anomaly generation is not to improve the model's predictions but to make the model more stable and robust to anomalies that may be encountered in the real world. Since no remarkable anomalies have been identified in the original dataset, the evaluation process we are about to carry out aims to empirically check that the predictions of the developed model are not devalued when artificial anomalies are introduced.

The prediction model employed for the tests is the same as in Section \ref{subsec:random_scen} based on feedforward neural networks with 2 hidden layers of 256 and 32 neurons, respectively, and an output layer with 2 neurons (one for each variable to be predicted). The objective of the model will also be to predict the next step of the multivariate time series. The amount of synthetic data generated is the same as in Section \ref{subsec:random_scen}, 15,000 samples with 10 scenarios each. Table \ref{tab:TSTR_Anomaly_Results} shows the results of the tests performed. In the tests where synthetic data and anomalies are used for training, the results show the mean of 5 tests performed and the confidence interval. We can observe that in the case of training only with synthetic data and anomalies (TSTR), the results are very close to the case of training the model only with real data (TRTR). In fact, adding anomalies even improves the TSTR results from the previous Table \ref{tab:TRTSresults}, where no anomalies are used. In the case of training with synthetic and real data together (TSRTR), adding anomalies still gives better results than training only with real data (TRTR). Although in this case, the results do not improve from those of the previous Table \ref{tab:TRTSresults}. 

Based on the tests performed, we can confirm with certain confidence that the addition of anomalies to the training dataset does not devalue the predictions made by the model and even improves them in some cases. Therefore, by integrating synthetic anomalies, we can achieve greater robustness and stability in the models without compromising their predictive performance.

\section{Conclusions and Future Directions}
\label{sec:conclusion}
This paper presents a methodology for the augmentation of time-series data from DCs using GANs. For this purpose, we have combined the approaches of data augmentation using GANs and scenario forecasting, allowing us to generate large amounts of synthetic data, taking advantage of the significant generation flexibility provided by GANs. The use of GANs also allows us to handle multivariate data with complex relationships and of different nature (e.g., categorical data). Furthermore, we propose a method to produce on-demand anomaly scenarios, increasing the synthetic data heterogeneity and the robustness of the models to be trained with these data. The production of anomalies does not require additional efforts and avoids endangering the electronic equipment's integrity. By including on-demand anomalous situations, the DC optimization algorithms will be able to cope more safely with unseen situations without compromising the physical integrity of electronic equipment while gathering these data.

To the best of our knowledge, our work is the first to propose a data augmentation methodology for DC facilities. Our research makes it possible to generate large volumes of synthetic data without compromising the security of DCs. To demonstrate our proposal's feasibility, we have used real data gathered from an operating DC, obtaining satisfactory results. However, our work can be easily adapted to any similar time-series-like problem in any field of application. These large volumes of synthetic data can improve AI-based optimization algorithms, enabling potential improvements in model optimization and energy consumption in DC facilities; thus, enabling a more sustainable and greener future. 

Future work lines include a more extensive hyperparameter search process and an empirical demonstration of our proposal's scalability to a larger number of variables. In DL literature, experiments have shown that providing additional information may improve the results and reliability. In the field of time series, we could add frequency spectrum information, predictions with classical methods (e.g., ARIMA), or time information (e.g., time, day of the week, time of the year). In this paper, we have also proposed two metrics to verify the realism of the generated data (KL Divergence and MSE). However, alternative metrics can be explored that, for example, contemplate the frequency spectrum. The field of generative models is one of the most active in the academic community today. Therefore, we believe it is essential that future research consider incorporating new state-of-the-art approaches, such as novel architectures and new training enhancements.

\section*{Acknowledgement}
This project has been partially supported by the Spanish Ministry of Science and Innovation under the grant PID2019-110866RB-I00, \textit{Adam Data Centers} and \textit{Tychetools}.

\bibliographystyle{plain}
\bibliography{sn-bibliography}


\end{document}